\documentclass[sigconf]{acmart}

\AtBeginDocument{}

\copyrightyear{2025}
\acmYear{2025}
\setcopyright{cc}
\setcctype{by}
\acmConference[SCF '25]{ACM Symposium on Computational Fabrication}{November 20--21, 2025}{Cambridge, MA, USA}
\acmDOI{XXXXXXX.XXXXXXX}
\acmISBN{978-1-4503-XXXX-X/2025/06}

\begin{document}

\title{AI Assisted AR Assembly: Object Recognition and Computer Vision for Augmented Reality Assisted Assembly}

% \title{AI-AR Assembly: Object and Gestural Recognition for Feedback-Assembly in Augmented Reality}

%% Authors and affiliations
\author{Alexander Htet Kyaw}
\affiliation{%
  \institution{Massachusetts Institute of Technology}
  \city{Cambridge}
  \state{MA}
  \country{United States}}
\email{alexkyaws@mit.edu}

\author{Haotian Ma}
\affiliation{%
  \institution{Cornell University}
  \city{Cornell}
  \state{NY}
  \country{United States}}
\email{hm443@mit.edu}

\author{Sasa Zivkovic}
\affiliation{%
  \institution{Cornell University}
  \city{Ithaca}
  \state{NY}
  \country{United States}}
\email{sz382@cornell.edu}

\author{Jenny Sabin}
\affiliation{%
  \institution{Cornell University}
  \city{Ithaca}
  \state{NY}
  \country{United States}}
\email{jes557@cornell.edu}

%% Short author list for headers
\renewcommand{\shortauthors}{Kyaw, Ma, Zivkovic, and Sabin}

%\begin{teaserfigure}
%  \label{Fig: Catalog}
%  \centering
%  \includegraphics[width=\linewidth]{figures/Hero-Thinner.jpg}
%  \caption{Self-Folding Chain Robot Reconfiguration Catalog}
%  \Description{Sequence of photos of the chain executing planar folds; some joints approach torque limits.}
%\end{teaserfigure}

\begin{abstract}
We present an AI-assisted Augmented Reality assembly workflow that uses deep learning-based object recognition to identify different assembly components and display step-by-step instructions. For each assembly step, the system displays a bounding box around the corresponding components in the physical space, and where the component should be placed. By connecting assembly instructions with the real-time location of relevant components, the system eliminates the need for manual searching, sorting, or labeling of different components before each assembly. To demonstrate the feasibility of using object recognition for AR-assisted assembly, we highlight a case study involving the assembly of LEGO sculptures. 

\end{abstract}

%% CCS Concepts (replace with generated terms when available)



\received{27 Sep 2025}

\maketitle

\section{Introduction}
By using Augmented Reality (AR), we can overlay digital information directly onto the physical world, providing 3D instructions throughout the assembly process. For example, prior work has explored combining AR assembly tasks with gesture recognition \cite{kyaw_gesture_2024}, computer vision \cite{jahn_depth_2022} and physics simulation to enhance interaction \cite{kyaw_active_2023}. 

Previous research has utilized AI algorithms such as object recognition to recognize assembly states, automatically advance AR instructions \cite{stanescu_state-aware_2023}, align virtual model registration \cite{canadinc_multi-3d-models_2024}, and verify whether a part has reached its target placement location \cite{kastner_integrative_2021}. 

Building upon this body of work, our system incorporates object recognition to detect individual parts, associate them with corresponding digital instructions, and dynamically highlight both their current and target spatial regions. This research advances the field of AR-assisted assembly by introducing a framework that delivers contextual, stepwise digital guidance aligned with physical components within the assembly space \cite{ma_ai_2023}.

\begin{figure} 
  \centering
  \includegraphics[width=1\linewidth]
   {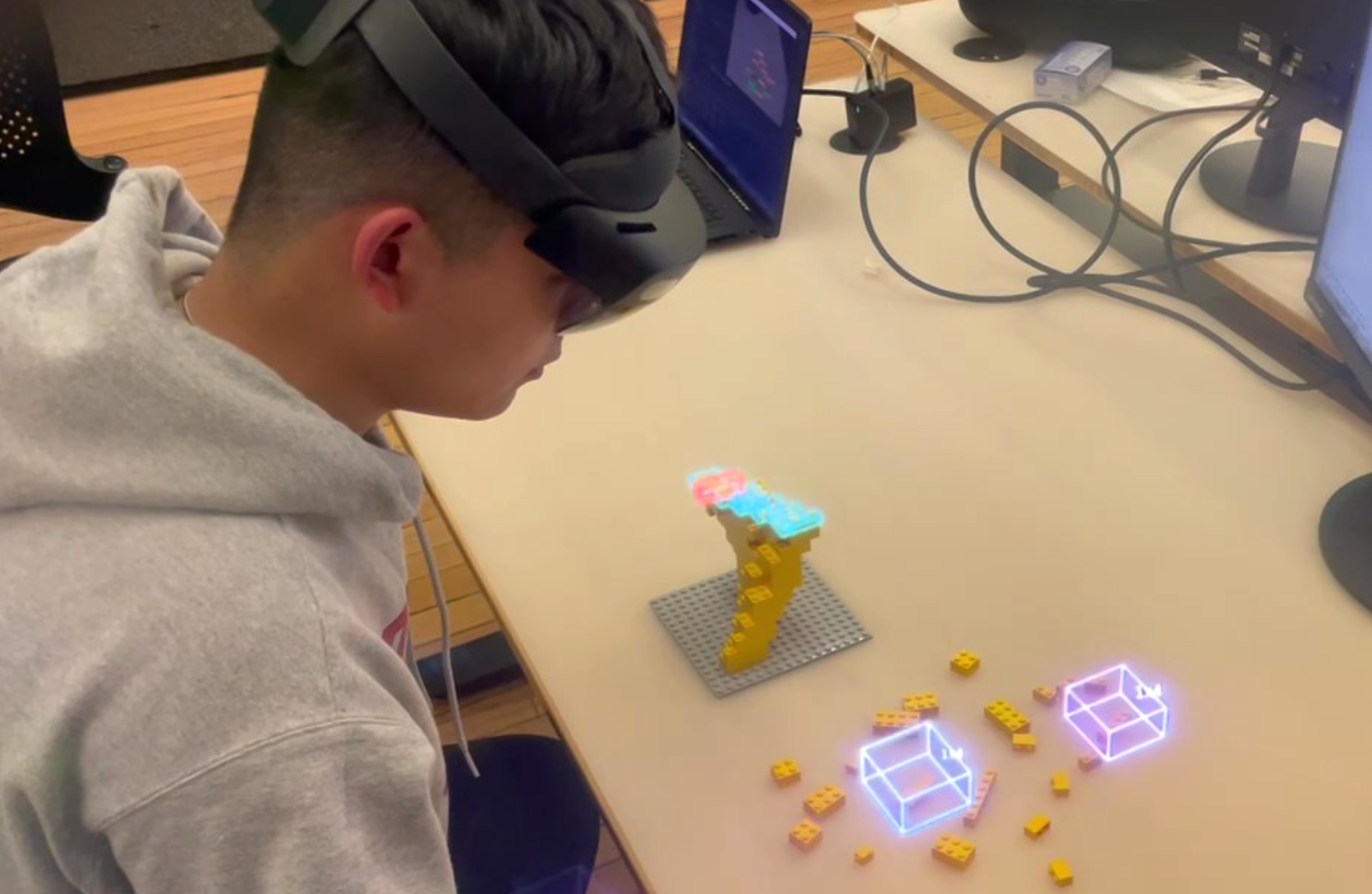}
   % {figures/3d_6-frame-sequence_03.jpg}
  \caption{Object Recognition for AR Assisted Assembly}
  \Description{Object Recognition for AR Assisted Assembly}
  \label{3D}
\end{figure}

\section{System Overview} 
Leveraging deep learning for object recognition, our system is capable of identifying each component within the assembly area and associating it with the relevant assembly instruction. The system highlights the necessary component for every assembly step by showing a bounding box around the component's current location and its intended placement location. It then automatically advances to the subsequent step once it detects that the current step has been completed. For the purpose of this research, the Microsoft Hololens 2 AR headset and Lego components were chosen as the case study to demonstrate the system's capabilities. 

\begin{figure*} 
  \centering
  \includegraphics[width=1\linewidth]
   {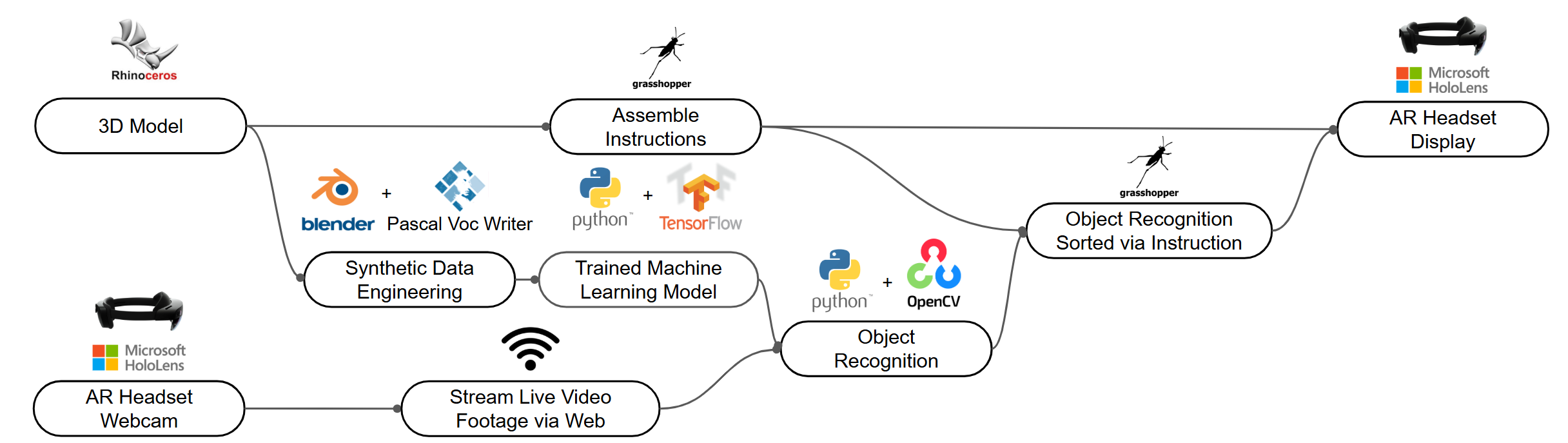}
   % {figures/3d_6-frame-sequence_03.jpg}
  \caption{System overview diagram illustrating the various software components and data -flow of the system}
  \Description{System overview diagram illustrating the various software components and dataflow of the system}
  \label{workflow}
\end{figure*}

\subsection{Object Recognition in Augmented Reality}
The object recognition algorithm is trained on synthetic data representing eight distinct primitive yellow LEGO components using the YOLOv5 model. These synthetic image datasets are generated using a rendering engine that simulates various orientations and lighting conditions to mimic real-world scenarios and improve detection robustness. The HoloLens 2 camera captures the physical workspace and streams video data to a server, where it is split into individual frames for object detection using the YOLOv5 model. The model identifies 2D bounding boxes around LEGO components in each frame. These bounding boxes are then projected into the AR environment using a homography-based 2D-to-3D planar projection, computed from the camera’s pose and field of view. See figure \ref{workflow} for software implementation details.

\begin{figure} [b]
  \centering
  \includegraphics[width=\linewidth]
   {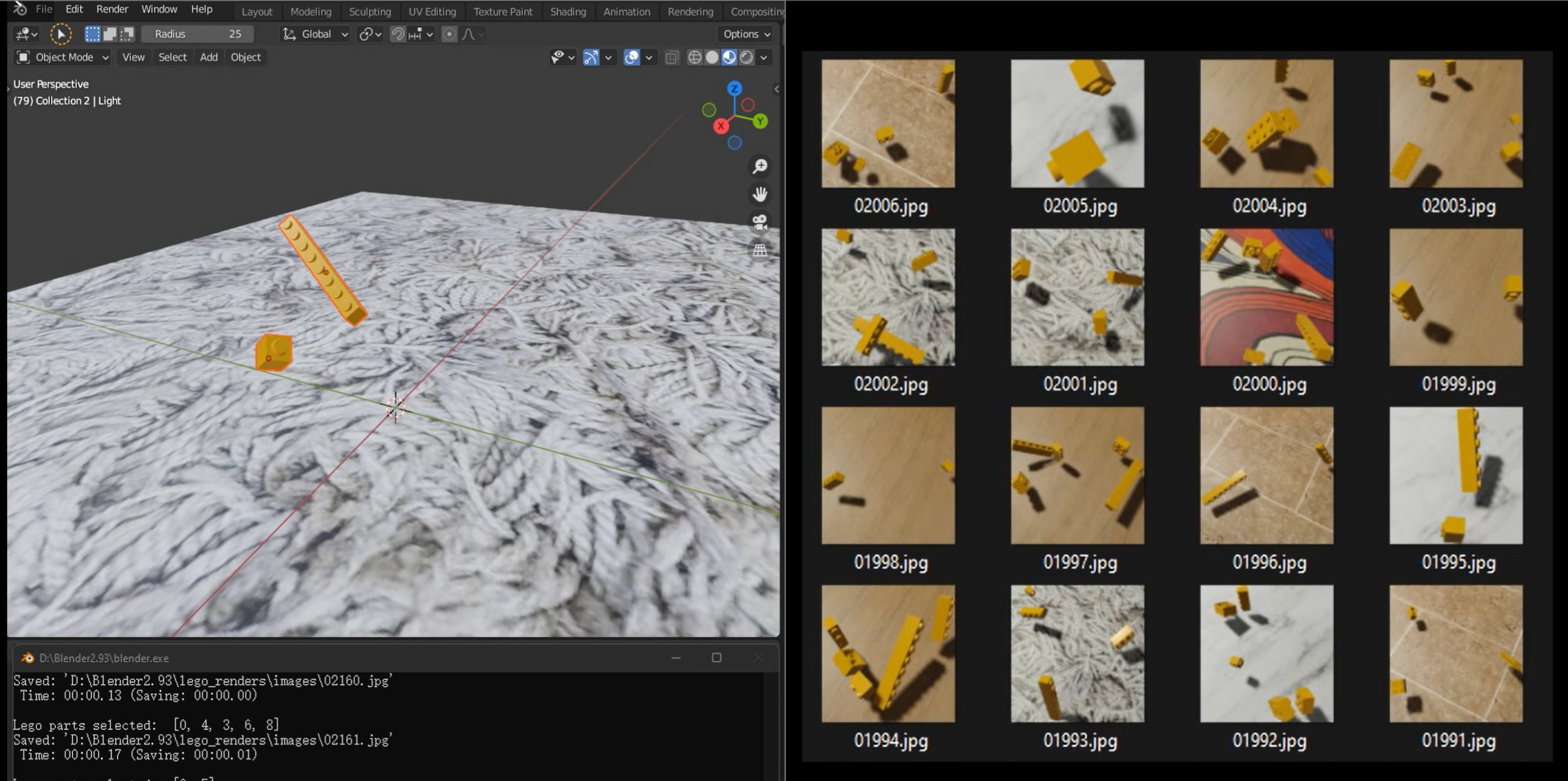}
   % {figures/3d_6-frame-sequence_03.jpg}
  \caption{Synthetic data generation for training object recognition algorithm}
  \Description{Synthetic data generation for training object recognition algorithm}
  \label{3D}
\end{figure}

\begin{figure} [b]
  \centering
  \includegraphics[width=\linewidth]
   {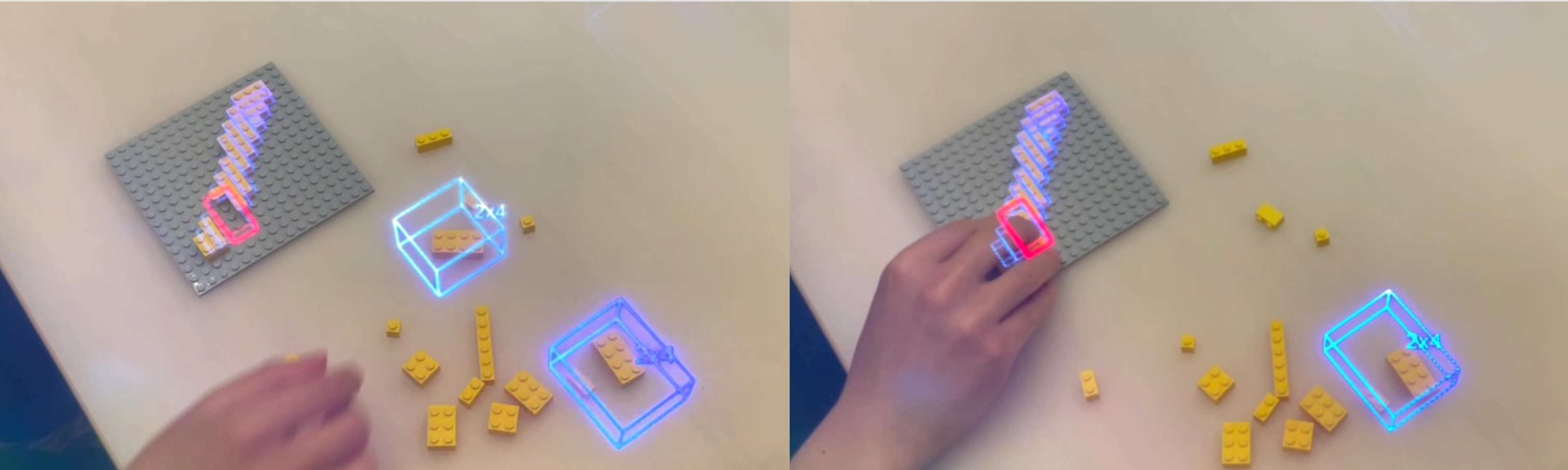}
   % {figures/3d_6-frame-sequence_03.jpg}
  \caption{Bounding Box Corresponding to Assembly Step}
  \Description{Bounding Box Corresponding to Assembly Step}
  \label{3D}
\end{figure}

\subsection{Step by Step 3D Assembly Instruction}
The interface displays step-by-step assembly instructions, indicating where the user should pick up each component and where to place it in relation to the rest of the assembly. For each instruction step, a corresponding 3D bounding box is generated to visually link the digital instruction with the physical component, highlighting both its current position and its intended placement. 3D bounding boxes are shown only for components relevant to the current assembly step, and each box is annotated with the component type, allowing users to identify both the location and the nature of the part. Since LEGO assemblies are built layer by layer, only the geometry of the current layer is visualized to reduce cognitive load. In our demonstration, the assembly parts are positioned on the right side of the workspace, while the assembly area is on the left.

%\begin{figure} [h]
%  \centering
%  \includegraphics[width=\linewidth]
%   {figures/steps.png}
%   % {figures/3d_6-frame-sequence_03.jpg}
%  \caption{3D folding sequences of various shapes}
%  \Description{Sequence of photos of the chain executing planar folds; some joints approach torque limits.}
%  \label{3D}
%\end{figure}

\section{Results}
We demonstrated the assembly of two distinct LEGO sculptures: an ellipsoidal egg and a twisted wall. Using our system, we successfully assembled both sculptures without referencing any 2D paper drawings or 3D digital models. Further user studies are planned to quantitatively evaluate the system’s effectiveness. While the current study is limited to tabletop assemblies, future research could explore more complex assembly tasks or improve the accuracy of the AR projections \cite{kyaw_augmented_2023}. Additionally, future work could explore using this system to assemble designs generated by 3D generative AI. \cite{kyaw_speech_2025,kyaw_making_2025}.
By bridging assembly instructions with the real-time localization of components in physical space, this research demonstrates the potential of object recognition for AI-assisted AR assembly.

% These are examples of mass-customized assemblies, meaning that the location and type of each component vary across configurations.

\begin{figure} [b]
  \centering
  \includegraphics[width=\linewidth]
   {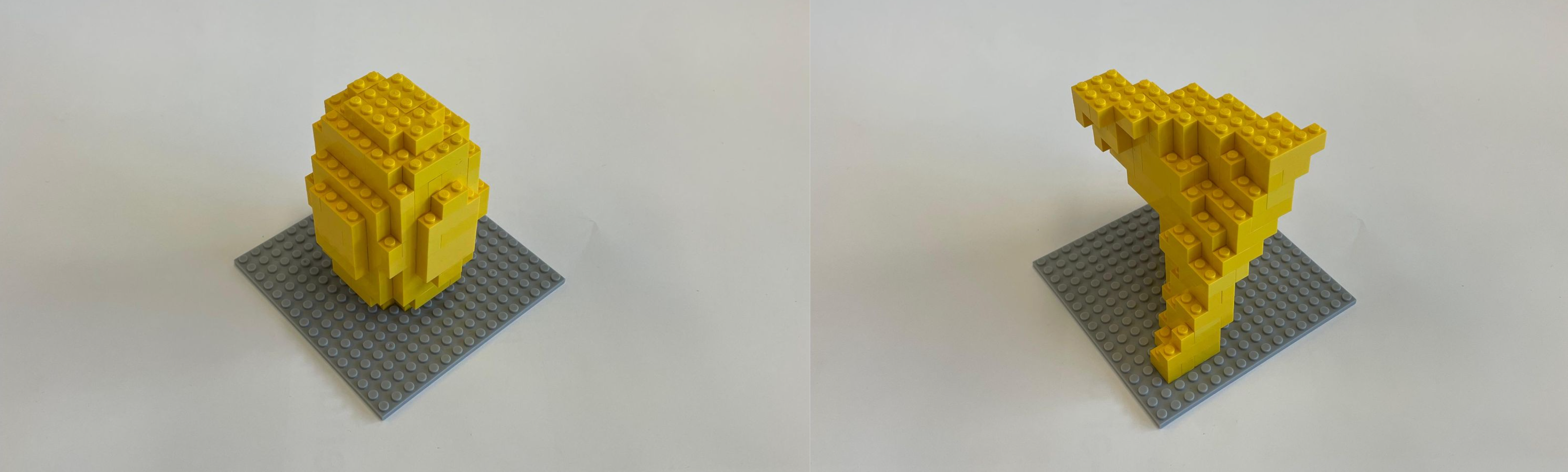}
   % {figures/3d_6-frame-sequence_03.jpg}
  \caption{Artifacts made using AI-Assisted AR Assembly}
  \Description{Artifacts made using AI-Assisted AR Assembly}
  \label{3D}
\end{figure}

\bibliographystyle{ACM-Reference-Format}
% Temporarily suppress BibTeX to avoid sample .bib warnings. Add your references file when ready.
% \bibliography{references}
\bibliography{references}

\end{document}